\RequirePackage{fix-cm}
\documentclass[twocolumn]{svjour3}          % twocolumn
\smartqed  % flush right qed marks, e.g. at end of proof
\usepackage{graphicx}
\usepackage{epsfig}
\usepackage{graphicx}
\usepackage[fleqn]{amsmath}
\usepackage[fleqn]{mathtools}
\usepackage{amssymb}
\usepackage{makecell}
\usepackage{multirow}
\usepackage{url}

 \journalname{International Journal of Computer Vision}

\newcommand{\etal}{\emph{et al. }}

\begin{document}

\title{Face Synthesis from Visual Attributes via Sketch using Conditional VAEs and GANs
}


\author{Xing Di         \and
        Vishal M. Patel
}


\institute{Xing Di \at
              Department of Electrical and Computer Engineering \\
              Rutgers, The State University of New Jersey\\
              94 Brett Road, Piscataway, NJ 08854\\
              Tel.: 848-445-5250\\
              \email{xd55@scarletmail.rutgers.edu}           
                         \and
           Vishal M. Patel \at
              Department of ECE, Rutgers University\\
               \email{vishal.m.patel@rutgers.edu}  
}

\date{Received: date / Accepted: date}

\maketitle

\begin{abstract}
Automatic synthesis of faces from visual attributes is an important problem in computer vision and has wide applications in law enforcement and entertainment.   With the advent of  deep generative convolutional neural networks (CNNs), attempts have been made to synthesize face images from attributes and text descriptions.  In this paper, we take a different approach, where we formulate the original problem as a stage-wise learning problem.   We first synthesize the facial sketch corresponding to the visual attributes and then we reconstruct the face image based on the synthesized sketch.   The proposed Attribute2Sketch2Face framework, which is based on a combination of  deep Conditional Variational Autoencoder (CVAE) and Generative Adversarial Networks (GANs), consists of three stages: (1) Synthesis of facial sketch from attributes using a CVAE architecture,  (2) Enhancement of coarse sketches to produce sharper sketches using a GAN-based framework, and (3) Synthesis of face from sketch using another GAN-based network.   Extensive experiments and comparison with recent methods are performed to verify the effectiveness of the proposed attribute-based three stage face synthesis method.

\end{abstract}

\section{Introduction}

Facial attributes are descriptions or labels that can be given to a face to describe its appearance \cite{kumar_ttributes}.  In the biometrics community, attributes are also referred to as soft-biometrics \cite{softbio}.  Various methods have been developed in the literature for predicting facial attributes from images \cite{DeepAtt}, \cite{kumar2008facetracer}, \cite{zhang2014panda}.   For instance, Kumar \etal \cite{kumar2008facetracer} proposed a facial part-based method for attribute predication.  Zhang \etal \cite{zhang2014panda} proposed a method which combines part-based models and deep learning for learning attributes.    Similarly, Liu \etal \cite{DeepAtt} proposed a convolutional neural network (CNN) based approach which combines two CNNs for localizing face region and extracting high-level features  from the localized region for predicting attributes.

\begin{figure}[t]
\centering
\includegraphics[width=1\linewidth]{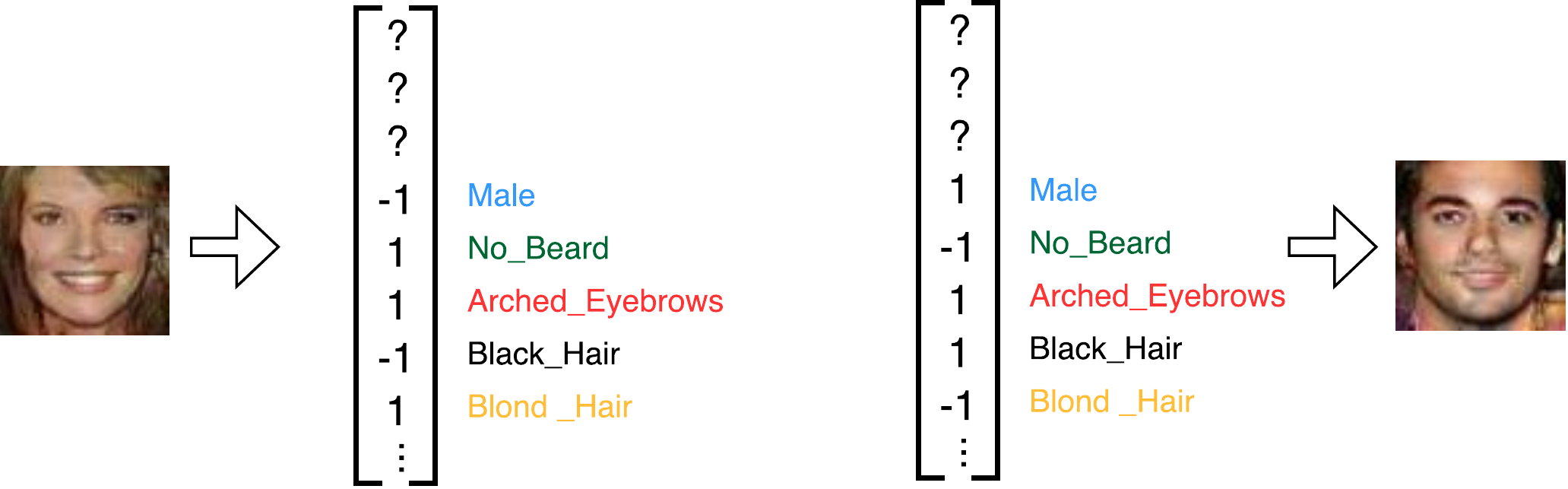}\\
(a)\hskip100pt (b)
\caption{Attribute prediction vs. face synthesis from attributes.  (a) Attribute prediction:  given a face image, the goal is to predict the corresponding attributes.  (b) Face synthesis from attributes: given a list of facial attributes, the goal is to generate a face image that satisfies these attributes.}
\label{fig:att_vs_rec}
\end{figure}

%\begin{figure*}[t]
%\centering
%\includegraphics[width=1\linewidth]{./fig/brief_architecture.pdf}\\
%\caption{Network architecture of the proposed Attribute2Sketch2Face method.  (a) Three-stage training network. (b) Testing phase. Here, $G2$ and $G3$ denote generators, while $D2$ and $D3$ denote discriminators.}
%\label{fig:overall_architecture}
%\end{figure*}

\begin{figure*}[t]
\centering
\includegraphics[width=1\linewidth]{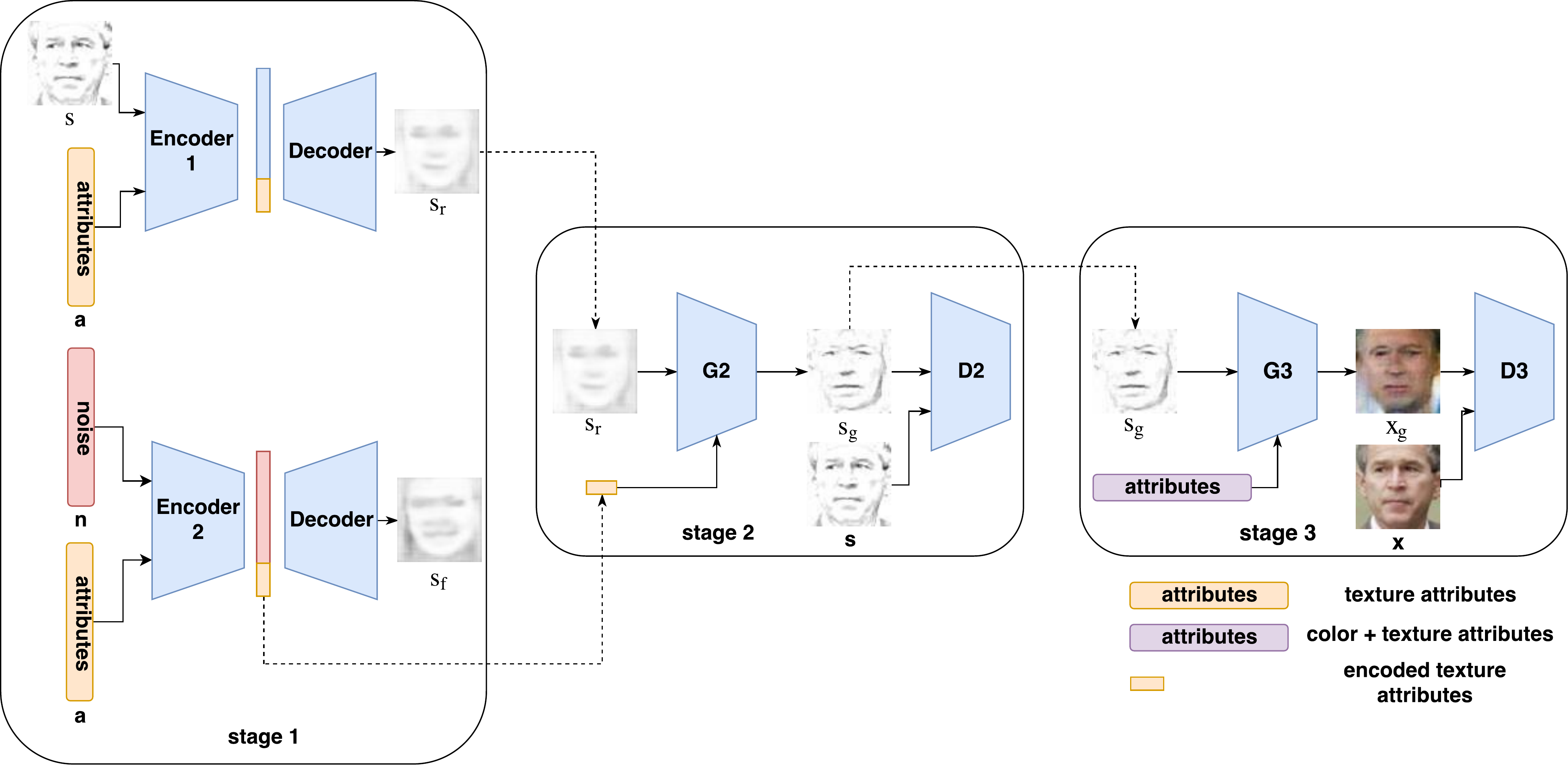}\\
\caption{Three-stage training network. Stage 1 generates a coarse approximation of the sketch image from attributes.  Stage 2 further enhances the sketch image from Stage 1.  Finally, Stage 3 generates a face image from the sketch generated from Stage 2 conditioned on the attributes.  Here, $G2$ and $G3$ denote generators, while $D2$ and $D3$ denote discriminators.  Note that the attributes are divided into two separate groups - one corresponding to texture and the other corresponding to color.}
\label{fig:training}
\end{figure*}

\begin{figure}[t]
\centering
\includegraphics[width=1\linewidth]{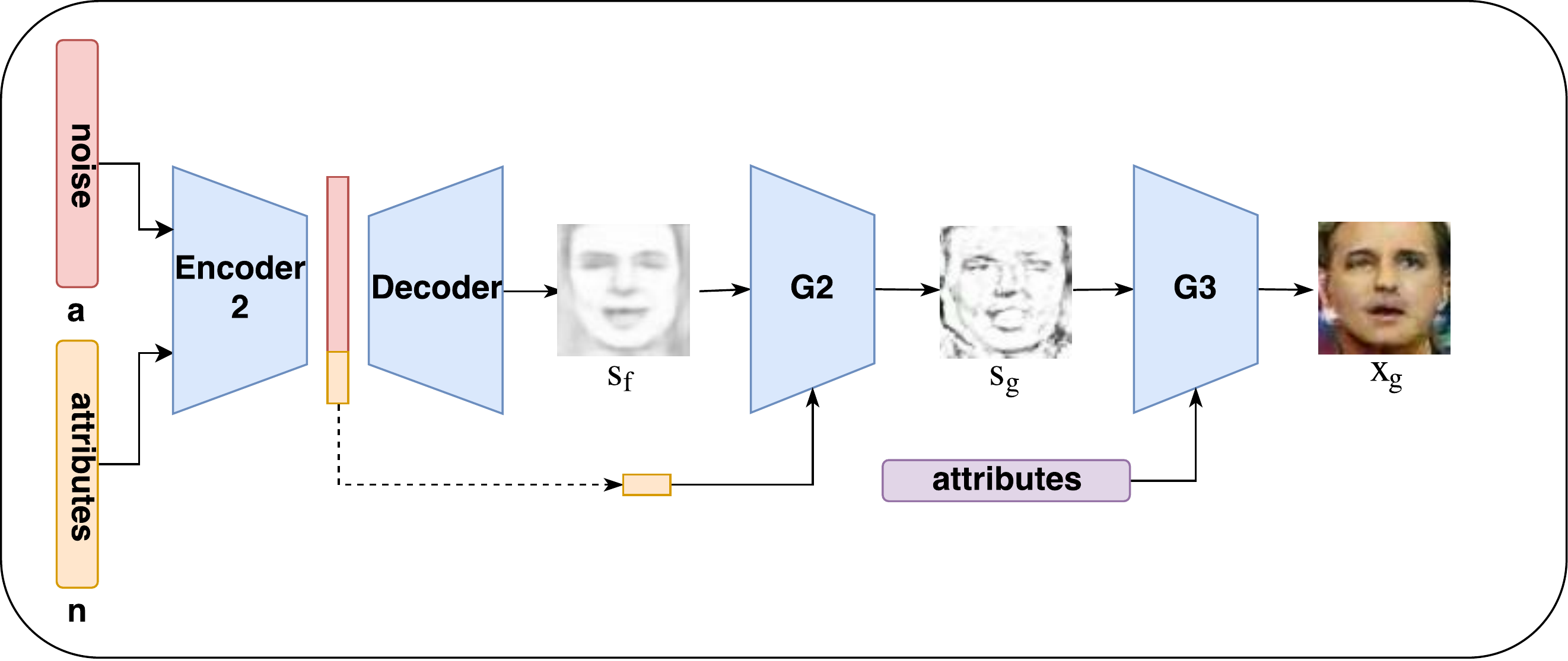}\\
\caption{Testing phase of the proposed Attribute2Sketch2Face method.  The network takes attributes and noise vectors as the input and generates high-quality face images.}
\label{fig:testing}
\end{figure}

While several methods have been proposed in the literature for inferring attributes from images, the inverse problem of synthesizing faces from their corresponding attributes is
a  relatively  unexplored  problem (see Figure~\ref{fig:att_vs_rec}). Visual description-based face synthesis has many applications in law enforcement and entertainment.  For example,  visual attributes are commonly used in law enforcement to assist in identifying suspects involved in a crime when no facial image of the suspect is available at the crime scene.  This is commonly done by constructing a composite or forensic sketch of the person based on the visual attributes.

Reconstructing an image from attributes or text descriptions is an extremely challenging problem.  Several recent works have attempted to solve this problem by using recently introduced CNN-based generative models such as conditional variational autoencoder (CVAE) \cite{sohn2015learning} and generative adversarial network (GAN) \cite{goodfellow2014generative}.  For instance, Yan \etal \cite{sohn2015learning} proposed a CVAE-based method for attribute-conditioned image generation.  In a different approach, Reed \etal \cite{reed2016generative} proposed a GAN-based method for synthesizing images from detailed text descriptions.  Similarly, Zhang \etal \cite{zhang2016stackgan} proposed a stacked GAN method for synthesizing photo-realistic images from text.

In contrast to the above mentioned methods, we propose a different approach to the problem of face image reconstruction from attributes.  Rather than directly reconstructing a face from attributes, we first synthesize a sketch image corresponding to the attributes and then reconstruct the face  image from the synthesized sketch.   Our approach is motivated by the way forensic sketch artists render the composite sketches of an unknown subject using a number of individually described parts and attributes.  

In particular, the proposed framework consists of three stages (see Figure~\ref{fig:training}).  In the first stage, we adapt a CVAE-based framework to generate a sketch image from visual attributes.  The generated sketch images from the first stage are often of poor quality.  Hence, in the second stage, we further enhance the sketch images using a GAN-based framework in which the generator sub-network leverages advantages of UNet \cite{ronneberger2015u} and DenseNet \cite{huang2017densely} architectures, which is inspired from \cite{jegou2017one}.  Finally, in the third stage, we reconstruct a color face image from the enhanced sketch image with the help of attributes using another GAN-based framework.  The Stage 3 formulation is motivated by the disentangled representation learning framework proposed in \cite{disentangled}.  In particular, the attribute information is fused with the latent representation vector to learn a disentangled representation.   Once the three-stage network is trained, one can synthesize sketches and face images by inputing visual attributes along with noise as shown in Figure~\ref{fig:testing}.  

To summarize, this paper makes the following contributions:
\begin{itemize}

\item We formulate the attribute-to-face reconstruction problem as a stage-wise learning problem (i.e. attribute-to-sketch, sketch-to-sketch, sketch-to-face).   

\item  A novel attribute-preserving dense UNet-based generator architecture, called AUDeNet, is proposed which incorporates the encoded texture attributes and the coarse sketches from stage 1 to generate  sharper sketches.  

\item  A new sketch-to-face synthesis generator is proposed which reconstructs the face image from the sketch image using attributes.  This generator is based on a new UNet structure and is able to preserve the attributes of the reconstructed image and improves the overall image  quality.

\item We use the combination of L1 loss, adversarial loss and perceptual loss \cite{johnson2016perceptual} in different stages for the purpose of image synthesis. 

\item  Extensive experiments  are conducted to demonstrate  the  effectiveness  of  the  proposed image synthesis method.  Furthermore, an ablation study is conducted to
demonstrate the improvements obtained by different stages of our framework.

\end{itemize}

 Rest of the paper is organized as follows.  In  Section~\ref{sec:related},
we review a few related works.  Details of the proposed attribute-to-face image synthesis method are given in Section~\ref{sec:method}.
Experimental results are presented in Section~\ref{sec:expt}, and finally, Section~\ref{sec:con} concludes the paper with a brief summary. 

Code is available at\\ \url{https://github.com/DetionDX/Attribute2Sketch2Face}.

\section{Background and Related Work} \label{sec:related}
Recent advances in deep learning have led to the development of various deep generative models for the problem of image synthesis and image-to-image translation \cite{larochelle2011neural}, \cite{kingma2013auto}, \cite{goodfellow2014generative}, \cite{rezende2014stochastic}, \cite{radford2015unsupervised}, \cite{sohn2015learning}, \cite{larsen2015autoencoding}, \cite{denton2015deep}, \cite{dosovitskiy2017learning}, \cite{salimans2016improved}, \cite{metz2016unrolled}, \cite{arjovsky2017towards}, \cite{che2016mode}, \cite{gauthier2014conditional}, \cite{odena2016conditional}.  Among them, variational autoencoder (VAE) \cite{kingma2013auto}, \cite{rezende2014stochastic}, GANs \cite{goodfellow2014generative}, \cite{radford2015unsupervised}, \cite{salimans2016improved}, and Autoregression \cite{larochelle2011neural} are the most widely used approaches.    

\subsection{Conditional VAE (CVAE)}
VAEs are powerful generative models that use deep networks to describe distribution of observed and latent variables. A VAE consists of two networks, with one network encoding a data sample to a latent representation and the other network decoding  latent representation back to data space. VAE regularizes the encoder by imposing a prior over the latent distribution. Conditional VAE (CVAE) \cite{sohn2015learning} \cite{yan2016attribute2image} is an extension  of VAE that models latent variables and data, both conditioned on side information such as a part or label of the image. The CVAE is trained by maximizing the variational lower bound
\begin{align}
\nonumber \mathcal{L}_{CVAE}(x,y;\theta,\phi) &= -KL(q_{\phi}(z|x,y)||p_{\theta}(z))\\&+\mathbb{E}_{q_{\phi}(z|x,y)}[\log p_{\theta}(x|y,z)],
\end{align}
where  $x, y$ and $z$ are input, output and latent variables, respectively, and $\theta$ and $\phi$ are the parameters.  Here, $p_{\theta}(z)$ is assumed to be an isotropic Gaussian distribution and $p_{\theta}(x|y,z)$ and $q_{\phi}(z|x,y)$ are multivariate Gaussian distributions.

\begin{figure*}[htp!]
\centering
\includegraphics[width=.85\linewidth]{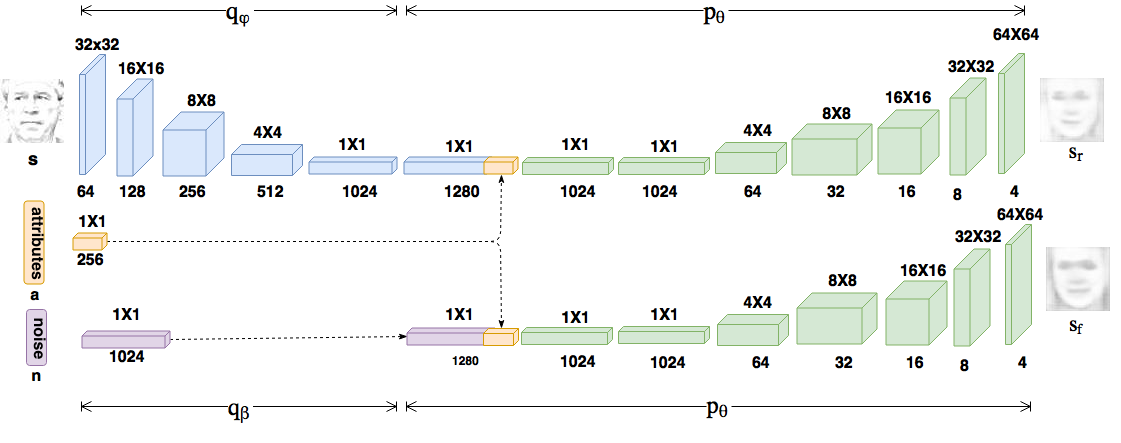}
\caption{Stage 1 (A2S) network architecture.} 
%We encode sketch, attribute and noise using an encoder $Q_{\theta}$ and $Q_{\beta}$ separately. After concatenate and reparameterization, we use a same decoder $P_{\phi}$ to project back to the sketch image.}
\label{fig:stage1_network}
\end{figure*}

\subsection{Conditional GAN}
GANs  \cite{goodfellow2014generative} are another class of generative models that are used to synthesize realistic images by effectively  learning the distribution of training images.  The goal of GAN is to train a generator, $G$, to produce samples from training distribution such that the synthesized samples are indistinguishable from actual distribution by the discriminator, $D$. Conditional GAN is another variant where the generator is conditioned on additional variables such as discrete labels, text  or images. 
The objective function of a conditional  GAN is defined as follows
\begin{equation}\label{eq:conditional GAN loss}
	\begin{split}
		L_{cGAN}(G,D) = E_{x,y \sim P_{data} (x,y)}[\log D(x,y)]+ \\
		E_{x\sim P_{data}(x),z\sim p_{z}(z)}[\log(1-D(x,G(x,z)))],
	\end{split}
\end{equation}
where $z$, the input noise, $y$, the output image, and $x$, the observed image, are sampled from distribution $P_{data} (x,y)$ and they are distinguished by the discriminator, $D$. While for the generated fake $G(x,z)$ sampled from distributions $x\sim P_{data}(x),z\sim p_{z}(z)$ would like to fool $D$.

Recently, several variants based on this game theoretic approach have been proposed for image synthesis and image-to-image translation tasks. Isola \etal \cite{isola2016image} proposed Conditional GANs \cite{mirza2014conditional} for several tasks such as labels to street scenes, labels to facades, image colorization, etc. In an another variant, Zhu \etal \cite{zhu2017unpaired} proposed CycleGAN that learns image-to-image translation in an unsupervised fashion. Berthelot \etal \cite{berthelot2017began} proposed a new method for training auto-encoder based GANs that is relatively more stable. Their method is paired with a loss inspired by Wasserstein distance \cite{arjovsky2017wasserstein}.   Reed \etal \cite{reed2016generative} proposed a conditional GAN network to generate reasonable images conditioned on the text description. Zhang \etal \cite{zhang2016stackgan} proposed a  two-stage stacked GAN method which achieves the state-of-art image synthesis results. Recently, Bao \etal \cite{bao2017cvae} proposed a fine-grained image generation method based on a combination of CVAE and GANs.  Yan \etal \cite{yan2016attribute2image} proposed a CVAE method using a disentangled representation in the latent and the original data distribution to achieve impressive attribute-to-image synthesis results.

Note that the approach we take in this paper is different from the above mentioned methods in that we make use of an intermediate representation (i.e. sketch) for the problem of image synthesis from attributes.  In contrast, some of the other methods attempt to directly reconstruct the image from attributes.   The only method that is closest to our approach is StackGAN \cite{zhang2016stackgan}, where the original image synthesis problem is broken into more manageable sub-problems through primitive shape and color refinement process.    Another important difference is that \cite{zhang2016stackgan} was specifically designed for text-to-image translation, while our approach is for the problem of attribute-to-face image reconstruction.    Furthermore, as will be shown later, our approach produces much better face reconstructions compared to \cite{zhang2016stackgan}.

\section{Proposed Method}\label{sec:method}
In this section, we provide details of the proposed Attribute2Sketch2Face method for image reconstruction from attributes.  It consists of three stages:  attribute-to-sketch (A2S), sketch-to-sketch (S2S), and sketch-to-face (S2F) (see Figure~\ref{fig:training}).  Note that the training phase of our method requires  ground truth attributes and the corresponding sketch and face images.   Furthermore, the attributes are divided into two separate groups - one corresponding to texture and the other corresponding to color.  Since sketch contains no color information, we use only texture attributes in A2S and S2S stages as indicated in Figure~\ref{fig:training}.

\subsection{Stage 1: Attribute-to-Sketch (A2S)}
In the A2S stage, we adapt the CVAE architecture from \cite{yan2016attribute2image}. Figure~\ref{fig:stage1_network} gives an overview of the Stage 1 network architecture. Given a texture attribute vector $a$, noise vector $n$, and ground-truth sketch $s$, we aim to learn a model $p_{\theta}(s|a, z)$ which can model the distribution of $s$ and generate  $s_{r}$. Here, $p_{\theta}$ denotes the decoder with parameter $\theta$ and
$z \sim q_{\phi}(z|s, a)$ denotes the encoder with parameter $\phi$. In this approach, the objective is to find the best parameter $\theta$ which maximizes the log-likelihood $\log p_{\theta}(s|a)$. In conditional VAE, the objective is to maximize the following variational lower bound, 
%\begin{align}\label{eq:cvae loglikelihood}
%\nonumber \log p_{\theta}(s|a) &= KL(q_{\phi}(z|s, a)||p_{\theta}(z|s, a))\\&+ \mathcal{L}_{CVAE}(s, a; \phi, \theta),
%\end{align}
%where the variational lower bound
\begin{align}
\nonumber \mathcal{L}_{CVAE}(s, a; \phi, \theta) &= -KL(q_{\phi}(z|s, a)||p_{\theta}(z))\\&+E_{z \sim q_{\phi}(z|s, a)}[\log p_{\theta}(s|a, z)],
\end{align}
where $p_{\theta}(z)$ is an isotropic multivariate Gaussian distribution and  $q_{\phi}(z|s, a)$ and $p_{\theta}(s|a, z)$ are two multivariate Gaussian distributions. The purpose of this function is to approximate the true conditional probability $p_{\theta}(s|a)$ with  $KL(q_{\phi}(z|s, a)||p_{\theta}(z|s, a))$ error by maximizing the  $\mathcal{L}_{CVAE}$ loss.

As shown in Figure~\ref{fig:stage1_network}, two encoders, $q_{\phi}$ and $q_{\beta}$ are the proposed Encoder 1 and Encoder 2, respectively in Figure~\ref{fig:training}.  The encoder $q_{\phi}$ takes sketch and attributes as input, whereas $q_{\beta}$ takes noise and attribute vectors as input. 
The overall loss function of the A2S stage is as follows
\begin{align}
 \label{stage 1 loss}
\nonumber \mathcal{L}_{A2S} &= \mathcal{L}_{CVAE}(s, a; \phi, \theta) - \lambda KL(q_{\beta}(z|n,a)||p_{\theta}(z))  
\\
\nonumber&= -KL(q_{\phi}(z|s,a)||p_{\theta}(z)) - \lambda KL(q_{\beta}(z|n, a)||p_{\theta}(z)) \\
&+ E_{z\sim q_{\phi}(z|s, a)}[\log p_{\theta}(s|a, z)],
\end{align}
The first two terms in \eqref{stage 1 loss},  $KL(q_{\phi}(z|s,a)||p_{\theta}(z))$ and $KL(q_{\beta}(z|n,a)||p_{\theta}(z))$, are the regularization terms in order to enforce the latent variable $z\sim q_{\phi}(z|s,a)$ and $z\sim q_{\beta}(z|n,a)$  both match the prior normal distribution, $p_{\theta}(z)$.

The encoder network  $q_{\phi}$ has two modules: one encoding the input sketch $s$ (in blue) and the other encoding the texture attribute $a$ (in yellow).  The encoding module for sketch $s$ consists of the following components: CONV5(64) - CONV5(128) - CONV3(256) - CONV3(512) - CONV4(1024),
  where CONVk(N) denotes N-channel convolutional layer with kernel of size $k\times k$. In particular, CONV5(64) and CONV5(128) consist of the convolutional layers followed by ReLU and 2-stride max pooling layer, respectively.  The next two layers CONV3(256) and CONV3(512) consists of the convolutional layers  followed by a batch normalization and ReLU layer, respectively.  The final CONV4(1024) layer consists of convolutional layers with kernel of size $4\times4$ with 1024-channel output. The other encoding module for attribute $a$ is a fully-connected network with 256-dimension output followed by 1D batch normalization and ReLU layers.

The encoder $q_{\beta}$, which takes the noise and attributes as input, also consists of
 the encoding module for attributes as in $q_{\phi}$ (shown in yellow) and the encoding module for noise (shown in purple). The noise encoding module consist of one fully-connected layer with 1024-dimensional output along with 1D batch normalization and ReLU layers. For the decoder $p_{\theta}(s|a, z)$ (shown in green), we first concatenate the encoded attributes with the encoded image/noise together and implement the reparameterization trick as in \cite{kingma2013auto}. Then reshape the mixed latent vector into a $4x4$ size feature maps. Then, we implement four UpsampleBlock which consists of 2D nearest upsampling layer followed by a $3\times 3$ convolutional layer, batch normalization and ReLU layers.

\begin{figure}[htp!]
\centering
\includegraphics[width=1\linewidth]{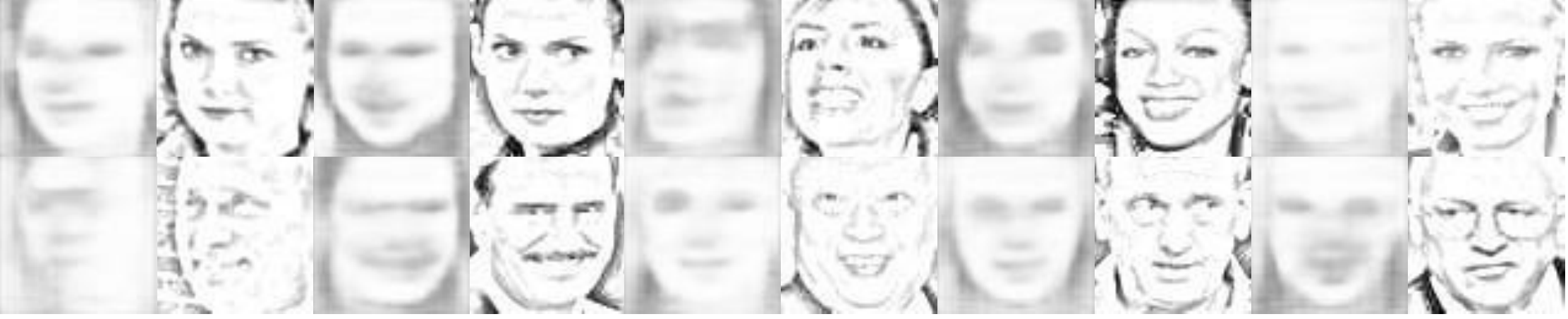}
\caption{Sample reconstruction results from Stage 1. Odd columns: reconstructed sketch images. Even columns: real sketch images.}
\label{fig:first_stage_result}
\end{figure}

\begin{figure}
\centering
\includegraphics[width=1\linewidth]{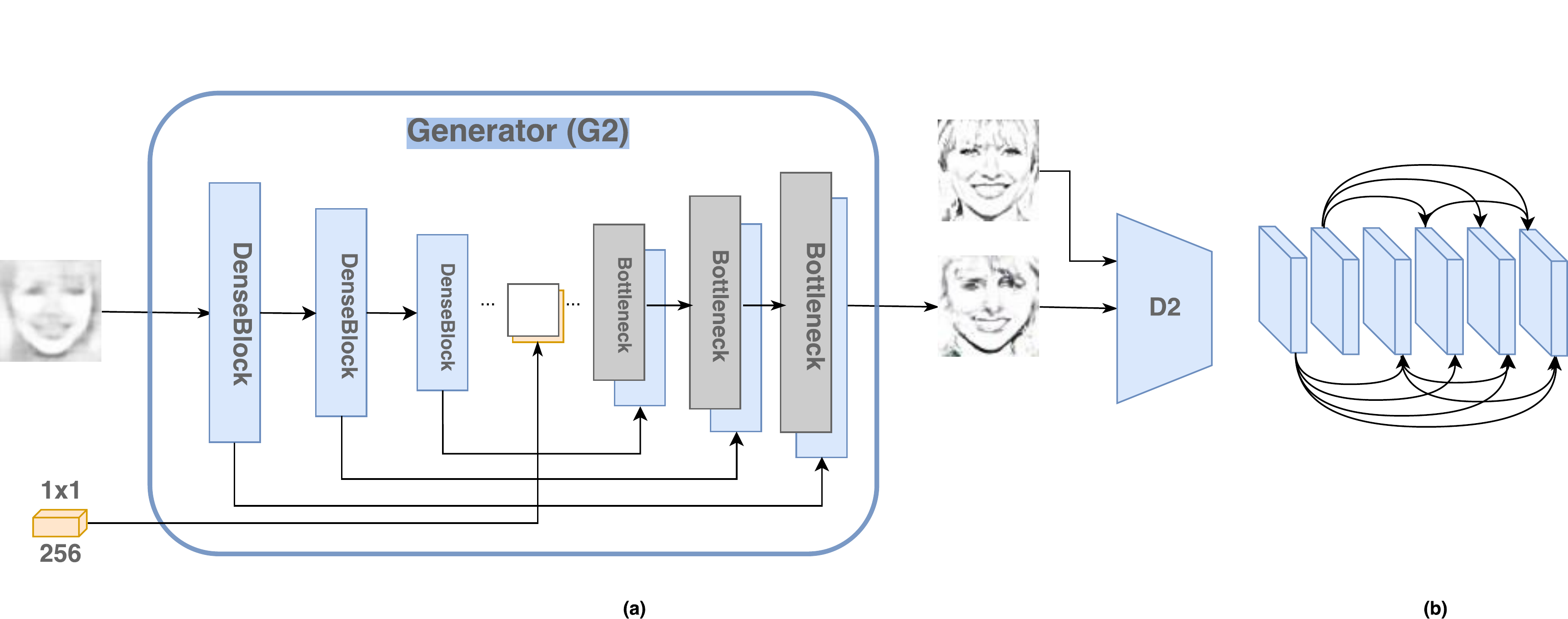}
\caption{Stage 2 (S2S) network architecture (AUDeNet).   (a) Generator ($G2$) produces sharp sketch images from blurry inputs.  Discriminator, ($D2$) is a patch-based discriminator with 4 downsampling blocks which is responsible to provide the adversarial feedback to $G2$. (b) The DenseBlock \cite{huang2016densely} used in $G2$.}
\label{fig:stage2_network}
\end{figure}

\subsection{Stage 2: Sketch-to-Sketch (S2S)}
As shown in Figure~\ref{fig:first_stage_result}, sketch reconstructions from Stage 1 are often of poor quality. Hence, we propose a conditional GAN-based framework  to generate sharper sketch images from blurry images.  As shown in Figure~\ref{fig:stage2_network}, the proposed network consists of a generator sub-network $G2$ (based on UNet \cite{ronneberger2015u} and DenseNet \cite{huang2017densely} architectures) conditioned on the encoded attribute vector from the A2S stage and a patch-based discriminator sub-network $D2$. $G2$ takes blurry sketch images as input and attempts to generate sharper sketch images, while $D2$ attempts to distinguish between real and generated images. The two sub-networks are trained iteratively.

\subsubsection{Generator (G2)}
Deeper networks are known to better capture high-level concepts, however, the vanishing gradient problem affects convergence rate as well as the quality of convergence. Several works have been developed to overcome this issue among which UNet \cite{ronneberger2015u} and DenseNet \cite{huang2017densely} are of particular interest. While UNet incorporates longer skip connections to preserve low-level features, DenseNet employs short range connections within micro-blocks resulting in maximum information flow between layers in addition to an efficient network. Motivated by these two methods, we propose AUDeNet for the generator sub-network $G2$ in which, the UNet architecture is seamlessly integrated into the DenseNet network in order to leverage advantages of both the methods. This novel combination enables more efficient learning and improved convergence quality.  Furthermore, in order to generate attribute preserving reconstructions, we concatenate the latent attribute vector from A2S with the latent vector from the encoder as shown in Figure~\ref{fig:stage2_network}.  

A set of 3 dense-blocks (along with transition blocks) are stacked in the front, followed by a set of 5 dense-block layers (transition blocks). The initial set of dense-blocks are composed of 6 bottleneck layers. For efficient training and better convergence, symmetric skip connections are involved into the  generator sub-network, similar to \cite{mao2016image}. Details regarding the number of channels for each convolutional layer are as follows: C(64) - M(64) - D(256) - T(128) - D(512) - T(256) - D(1024) - T(512) - D(1024) - DT(256) - D(512) - DT(128) - D(256) - DT(64) - D(64) - D(32) - D(32) - DT(16) - C(3), where C(K) is a set of $K$-channel convolutional layers followed by batch normalization and ReLU activation. M is max-pooling layer. D(K) is the dense-block layer with $K$-channel output, T(K) is transition layer with $K$-channel output for downsampling. DT(K) is similar to T(K) except for transposed convolutional layer instead of convolutional layer for upsampling. 

\subsubsection{Discriminator (D2)}
Motivated by \cite{isola2016image}, patch-based discriminator $D2$ is used and it is trained iteratively along with $G2$. The primary goal of $D$ is to learn to discriminate between real and synthesized samples. This information is backpropagated into $G$ so that it generates samples that are as realistic as possible. Additionally, patch-based discriminator ensures preserving of high-frequency details which are usually lost when only L\textsc{1} loss is used. All the convolutional layers in $D2$ have a filter size of $4\times4$.  

\subsubsection{Objective function}
The network parameters for the S2S stage are learned by minimizing the following objective function:  
\begin{equation}\label{eq:overall_loss}
\mathcal{L} = \mathcal{L}_{A} + \lambda_{1} \mathcal{L}_{1} +\lambda_{2}\mathcal{L}_{perp},
\end{equation}
where $\mathcal{L}_{A}$ is the adversarial loss, $\mathcal{L}_{1}$ is the loss based on the $L_1$-norm between the synthesized image and the target, $\mathcal{L}_{perp}$ is the perceptual loss, and  $\lambda_{1}, \lambda_{2}$ are weights.  Adversarial loss is based primarily on the discriminator sub-network $D2$. Given a set of $N$ synthesized sketch images, $\{s_{g}\}_{i=1}^{N}$, the entropy loss from $D2$ that is used to learn the parameters of $G2$ is defined as $$\mathcal{L}_{A} = -\frac{1}{N}\sum_{i=1}^{N}\log(D2(s_{g})).$$
The L\textsc{1} loss measures the reconstruction error between  the synthesized sketch image and the corresponding target sketch and is defined as $$\mathcal{L}_{1} = \|s_{g}-s\|_{1}.$$
Finally, the perceptual loss \cite{johnson2016perceptual} is used to measure the distance between high-level features extracted  from  a  pre-trained CNN  and is defined as $$\mathcal{L}_{perp} = \|V(s_{g})-V(s)\|_{1}.$$
Here,  $s$ and $s_{g}$ indicate target and synthesised images respectively and $V$ is a particular layer of the VGG-16 network.  In our work, the output from the conv1-2 layer of a pre-trained VGG-16 network \cite{simonyan2014very} is used as the feature representation. Note that, the coarse sketches from the previous Stage 1, along with the corresponding target sketches, are used to train the network.

\subsection{Stage 3: Sketch-to-Face (S2F)}
\begin{figure}
\centering
\includegraphics[width=1\linewidth]{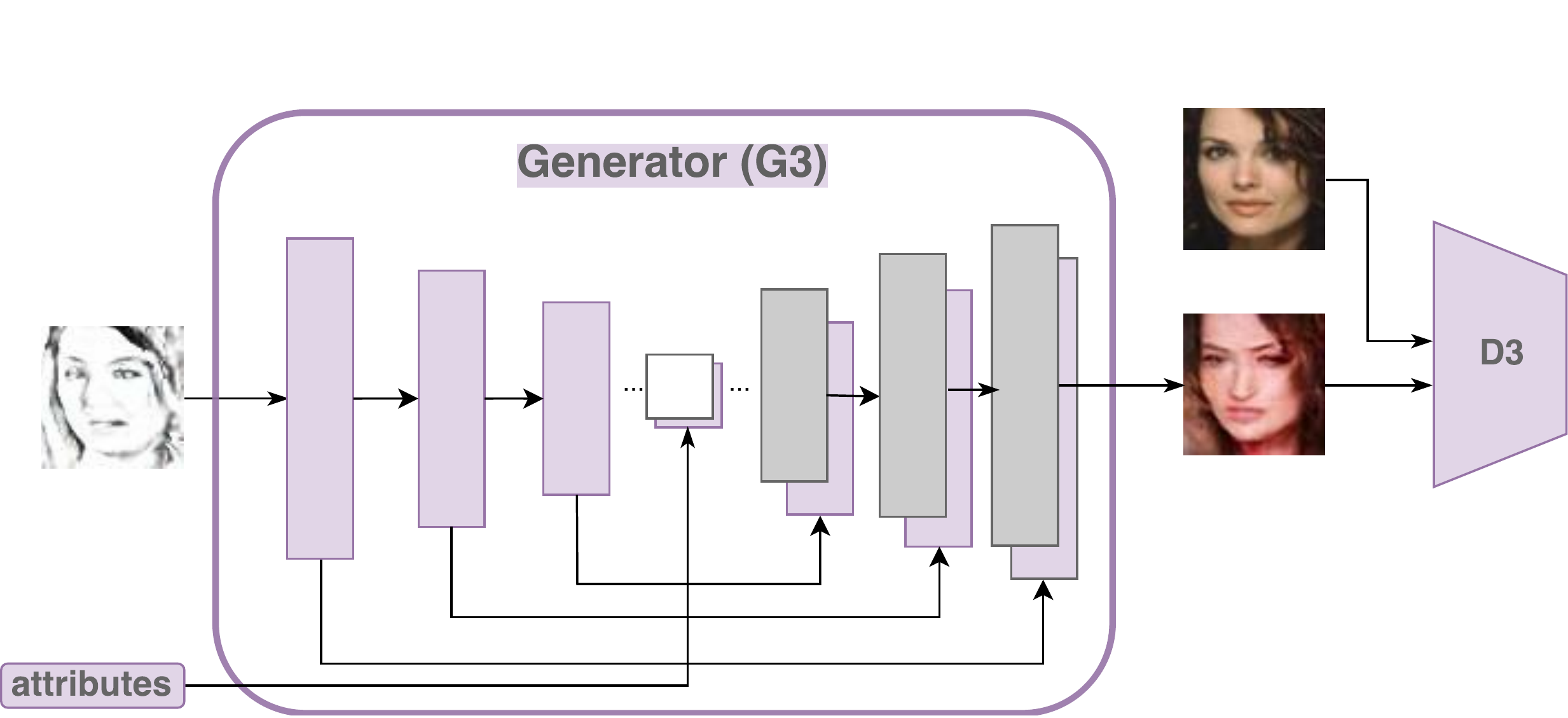}
\caption{Stage 3 (S2F) network architecture.  A novel UNet-based generator, $G3$, conditioned on visual attributes is used to synthesize face images from the sketch images.  $D3$ is a patch-based discriminator.}
\label{fig:stage3_network}
\end{figure}

The objective of Stage 3 is to reconstruct a color face image from the sketch image generated from the S2S stage.  We propose a GAN-based framework for this problem where we make use of another UNet-based architecture for the generator sub-network.  In particular, the visual attribute vector is combined with the latent representation to produce attribute-preserved image reconstructions.  Figure~\ref{fig:stage3_network} gives an overview of the proposed network architecture for S2F.     

\subsubsection{Generator (G3)}
The Stage 3 generator consists of five convolutional layers and five transposed convolutional layers.  Details regarding the number of channels for each convolutional and transposed convolutional layers are as follows:
C(64) - C(128) - C(256) - C(512) - C(512) - R(512) - DC(512) - DC(256) - DC(128) - DC(64) - DC(1), where C(K) is a set of $K$-channel convolutional layers followed by batch normalization and leaky ReLU activation.  DC(K) denotes a set of $K$-channel transposed convolutional layers along with ReLU and batch normalization layers.  R(C) is a two-layer ResNet Block as in StackGAN \cite{zhang2016stackgan} to fuse the attribute vector with the UNet latent vector.  Note that unlike Stages 1 and 2, the attribute vector here consists of both texture and color attributes.

\subsubsection{Discriminator (D3)}
Similar to $D2$, a patch-based discriminator $D3$, consisting of 4 downsampling blocks, is used and it is trained iteratively along with $G3$.

\subsubsection{Objective function}
The network parameters for the S2F stage are learned by minimizing \eqref{eq:overall_loss}.  In particular, a combination of $L_{1}$ loss, adversarial loss and perceptual loss is used.  As before, the perceptual loss is measured by using the deep feature representations from the conv1-2 layer of a pre-trained VGG-16 network \cite{simonyan2014very}.  We use the enhanced sketch from the previous stage along with the target face image to train this network.

\subsection{Testing}
Figure~\ref{fig:testing} shows the testing phase of the proposed method.  Attribute and noise vectors are first passed through the encoder/decoder structure corresponding to the A2S stage.  The encoded texture attribute vector along with the generated sketch from the A2S stage are fed into an AUDeNet-based generator (G2) to produce a sharper sketch image.  Finally, a UNet-based attribute-conditioned generator (G3) corresponding to the S2F stage is used to reconstruct a high-quality face image from the sketch image generated from the S2S stage.  In other words, our method takes noise and attribute vectors as input and generates high-quality face images via sketch images.

\section{Experimental Results} \label{sec:expt}
In  this  section,  experimental  settings  and  evaluation  of
the  proposed  method  are  discussed  in  detail.  Results  are  compared  with  several
state-of-the-art  generative  models:  CVAE \cite{sohn2015learning} adapted from \cite{yan2016attribute2image}, text2img \cite{reed2016generative} and stackGAN \cite{zhang2016stackgan}.  In addition, we compare the performance of our method with a baseline, attr2face, in which we attempt to recover the image directly from attributes without going to the intermediate stage of sketch.   The entire network in Figure~\ref{fig:training} is trained  stage-by-stage using Pytorch \footnote{https://github.com/pytorch/pytorch}.

\subsection{Datasets}
\begin{figure}[htp!]
\centering
\includegraphics[width=1\linewidth]{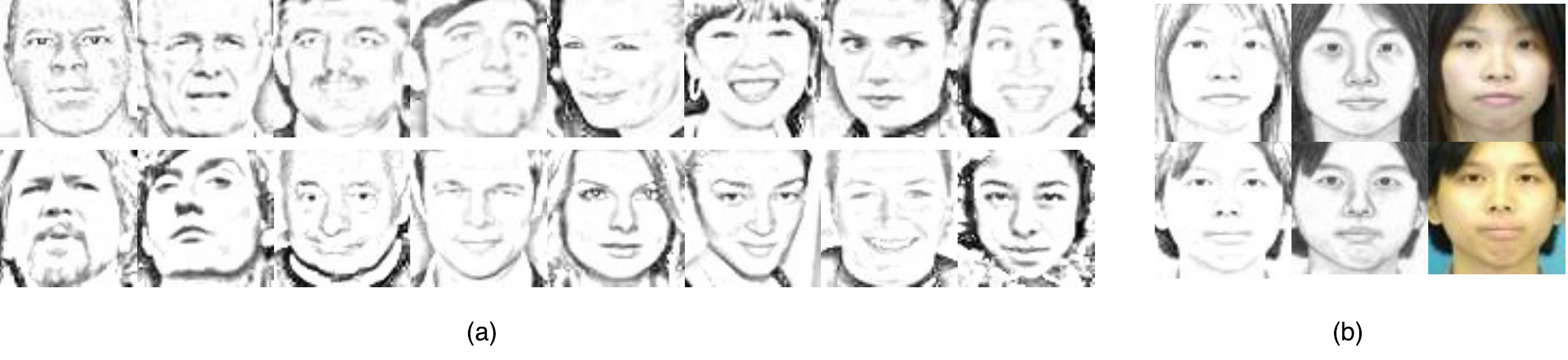}
\caption{Generated sketch images. (a) Sketch images from the LFW and the CelebA datasets are shown in row 1 and row 2, respectively. (b) Left to right: Comparison of the composed sketch, real sketch and real photo from the CUHK dataset. As can be seen from this figure that the composed sketch images are very similar to the ones drawn by artists and they preserve the texture and shading information present in the color images.  Hence, they can be used as a good replacement for real sketches.}
\label{fig:sketch_example}
\end{figure}

We conduct experiments using three publicly available datasets: CelebA \cite{liu2015faceattributes}, deep funneled LFW \cite{Huang2012a} and CUHK  \cite{wang2009face}.   The CelebA database contains about 202,599 face images, 10,177 different identities and 40 binary attributes for each face image.  The deep funneled LFW database contains about 13,233 images, 5,749 different identities and 40 binary attributes for each face image which are from the LFWA dataset \cite{liu2015faceattributes}. The CUFS dataset \cite{wang2009face} consists of 88 real sketches and photos for training, and 100 real sketches and photos for testing.   For   each   face image in the CUHK dataset,   the corresponding sketch image was  drawn  by  an  artist  when  viewing  this  photo.  Note that the training part of our network requires original face images and the corresponding sketch images as well as the corresponding list of visual attributes.  The CelebA and the deep funneled LFW datasets consist of both the original images and the corresponding attributes while the CUHK dataset consists of face-sketch image pairs.   To generate the missing sketch images in the CelebA and the deep funneled LFW datasets, we use a pencil-sketch synthesis method \footnote{http://www.askaswiss.com/2016/01/how-to-create-pencil-sketch-opencv-python.html} to generate the sketch images from the face images.   The missing attributes in the CUHK dataset were manually labeled.  Figure~\ref{fig:sketch_example}(a) shows some sample generated sketch images from  the CelebA and the deep funneled LFW datasets.  Figure~\ref{fig:sketch_example}(b)  shows the synthetic sketches, real sketches and real face images examples from CUHK.

\subsection{Preprocessing}
The MTCNN method \cite{zhang2016joint} was used to detect and crop faces from the original images.  The detected faces were rescaled to the size of $64\times 64$.  Since many attributes from the original list of 40 attributes were not significantly informative, we selected 23 most useful attributes for our problem.  Furthermore, the selected attributes were further divided into 17 texture and 6 color attributes as shown in Table~\ref{tab:fine-grained attributes}.   During experiments, the texture attributes were used for generating sketches in the A2S and S2S stages while all 23 attributes were used for generating high-quality face images in the final S2F stage.  

\begin{table}[htp!]
\centering
\caption{List of fine-grained texture and color attributes.}\label{tab:fine-grained attributes} 
\begin{tabular}{|c|c|}
\hline  Texture &  \makecell{Arched\_Eyebrows, Bags\_Under\_Eyes, Bald,\\ Bangs, Big\_Lips, Big\_Nose,\\ Bushy\_Eyebrows, Chubby,\\ Male, Narrow\_Eyes, No\_Beard,\\ Smiling, Young} \\ 
\hline  Color& \makecell{Black\_Hair, Blond\_Hair, Brown\_Hair, \\Gray\_Hair, Pale\_Skin, Rosy\_Cheeks} \\ 
\hline 
\end{tabular}
\end{table}

\begin{figure*}[htp!]
\centering
\includegraphics[width=1\linewidth]{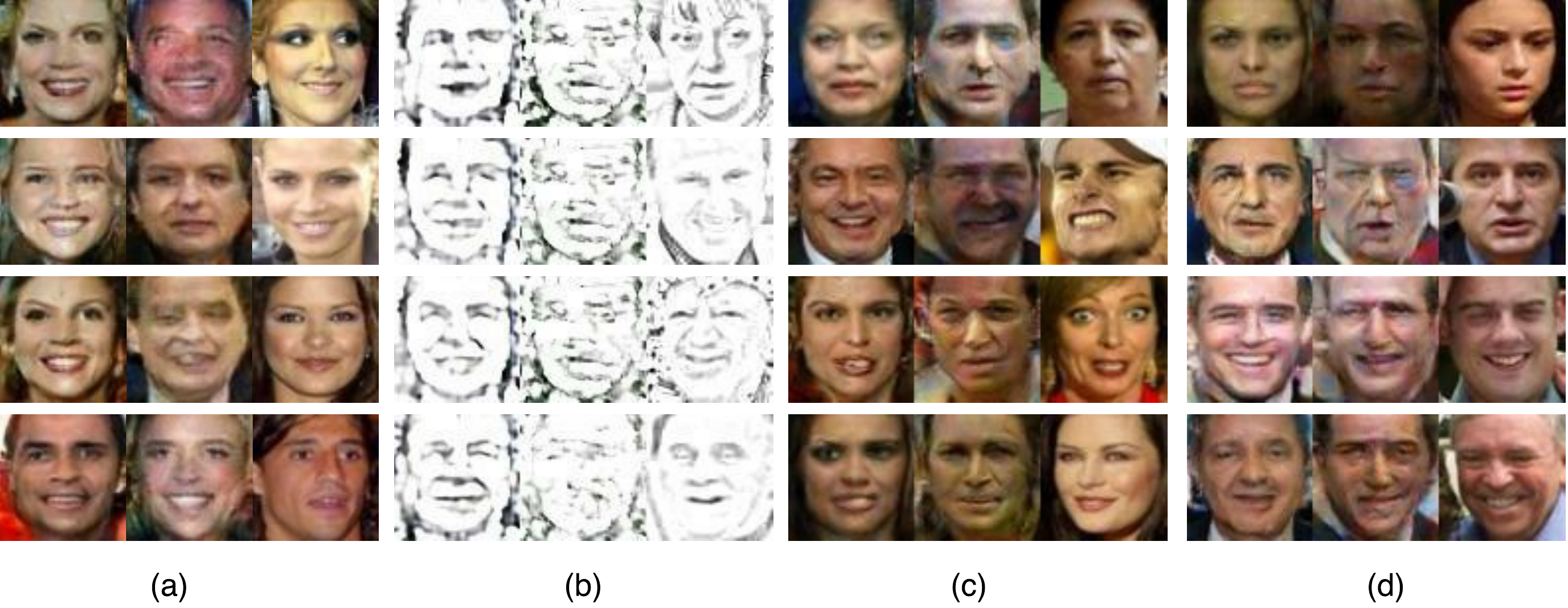}
\caption{Comparison of results from different configurations of the proposed network.  For all  subfigures (a) (b) (c) and (d):  first column: output using the proposed method,  second column: output with the specific configuration, and third column: reference images. (a)  Results corresponding to the case where attributes are not used in Stage 3. (b) S2S reconstructions without the use of encoded texture attributes.  (c) Results when the second stage of S2S is omitted from the pipeline.  Results show reconstructions with wrong attributes.  (d) Results when the second stage of S2S is omitted from the pipeline.  Poor quality reconstructions are obtained when S2S is skipped from the proposed method.}
\label{fig:ablation_study}
\end{figure*}

\subsection{Ablation Study}
In this section, we perform an ablation study to demonstrate  the  effects  of  different  modules  in  the  proposed
method. The following three configurations are evaluated.
\begin{enumerate}
\item Omit attributes while enhancing the sketch images generated from the A2S stage.  This will show the significance of using attributes while enhancing the sketch images in the S2S stage.
\item Remove the second stage of sketch image enhancement from the entire pipeline.  In other words, reconstruct the face image directly from the blurry sketch generated from A2S without enhancement.   This will clearly show the significance of the S2S stage.    
\item Remove the attribute concatenation from the final S2F stage.  This will show the significance of using 
 attributes in the final stage of sketch-to-image generation. 
\end{enumerate}
Results corresponding to the above three configurations are shown in Figure~\ref{fig:ablation_study}.
Results corresponding to the first experiment are shown in Figure~\ref{fig:ablation_study}(b), where the first, second and third columns indicate, the outputs from the S2S stage of our method, reconstructions without the use of attributes in S2S, the reference sketch, respectively. From this figure we clearly see that attribute-conditioned generator $G2$ produces sketches that are much better than the ones where sketches are enhanced directly without conditioning on the attributes.  
  
Results corresponding to the third experiment are shown in Figure~\ref{fig:ablation_study}(a), where the first, second and third columns show the reconstruction results from our method, reconstructions without using attributes in S2F, and reference images, respectively.  As can be seen from this figure, the absence of attributes in the final stage results in reconstructions with wrong face features such as gender and hair.  When attributes are used along with the sketch from S2S, the produced results have attributes that are very close to the ones corresponding to the original images.  This can be clearly seen by comparing the first and last columns in Figure~\ref{fig:ablation_study}(a).

In the final experiment, we omit the second stage of S2S from our pipeline and attempt to reconstruct the image from attributes in a two-stage procedure.  In other words, sketch images generated from the A2S stage are directly fed into the S2F stage.    Results are shown in Figure~\ref{fig:ablation_study}(c) and (d).  In both figures, first, middle and last columns show reconstructions from our method, without the second stage and reference images, respectively.    As can be seen from these figures, omission of the S2S stage from our pipeline produces images that are of poor quality (see results in Figure~\ref{fig:ablation_study}(d)).  The enhancement of sketches in Stage 2 not only produces sharper results but also with correct attributes (see results in Figure~\ref{fig:ablation_study}(c)).

\subsection{CelebA Dataset Results}
\begin{figure*}[htp!]
\centering
\includegraphics[width=.8\linewidth]{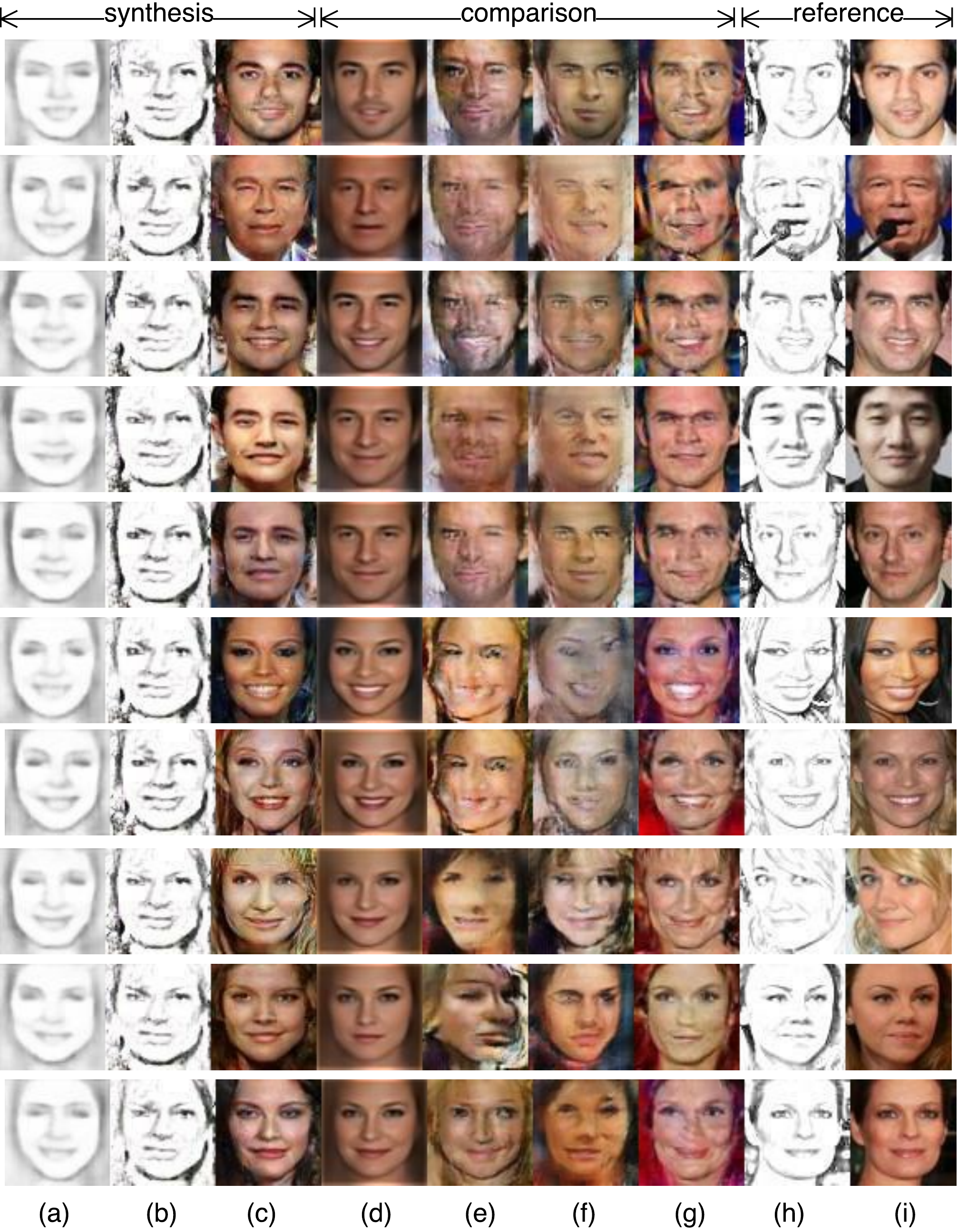}
\caption{Image reconstruction results on the CelebA dataset. (a)  Stage 1 results, (b)  Stage 2 results, (c) Stage 3 results (output from our method), (d) CVAE \cite{yan2016attribute2image}, (e) text2img \cite{reed2016generative}, (f) StackGAN \cite{zhang2016stackgan}, (g) attr2face, (h) reference sketch, (i) reference face image.}
\label{fig:CelebA_result}
\end{figure*}

The CelebA dataset \cite{liu2015faceattributes} consists of 162,770 training samples, 19,867 validation samples and 19,962 test samples.  After preprocessing and combining the training and validation sets, we obtain 182,468 samples which we use for training our three-stage network.   After preprocessing, the number of samples in the test set remain the same.  During training, we used a batch size of 128.  The ADAM algorithm \cite{adam_opt} with learning rate of 0.0002 is used.   We keep this initial learning rate for the first 10 epochs.  For the next 10 epochs, we let it drop by 1/decay\_epoch of its previous value after every epoch which is 1/10.  The total training time was about 20 hours in a single Titan X GPU.

 \begin{figure*}[htp!]
\centering
\includegraphics[width=.8\linewidth]{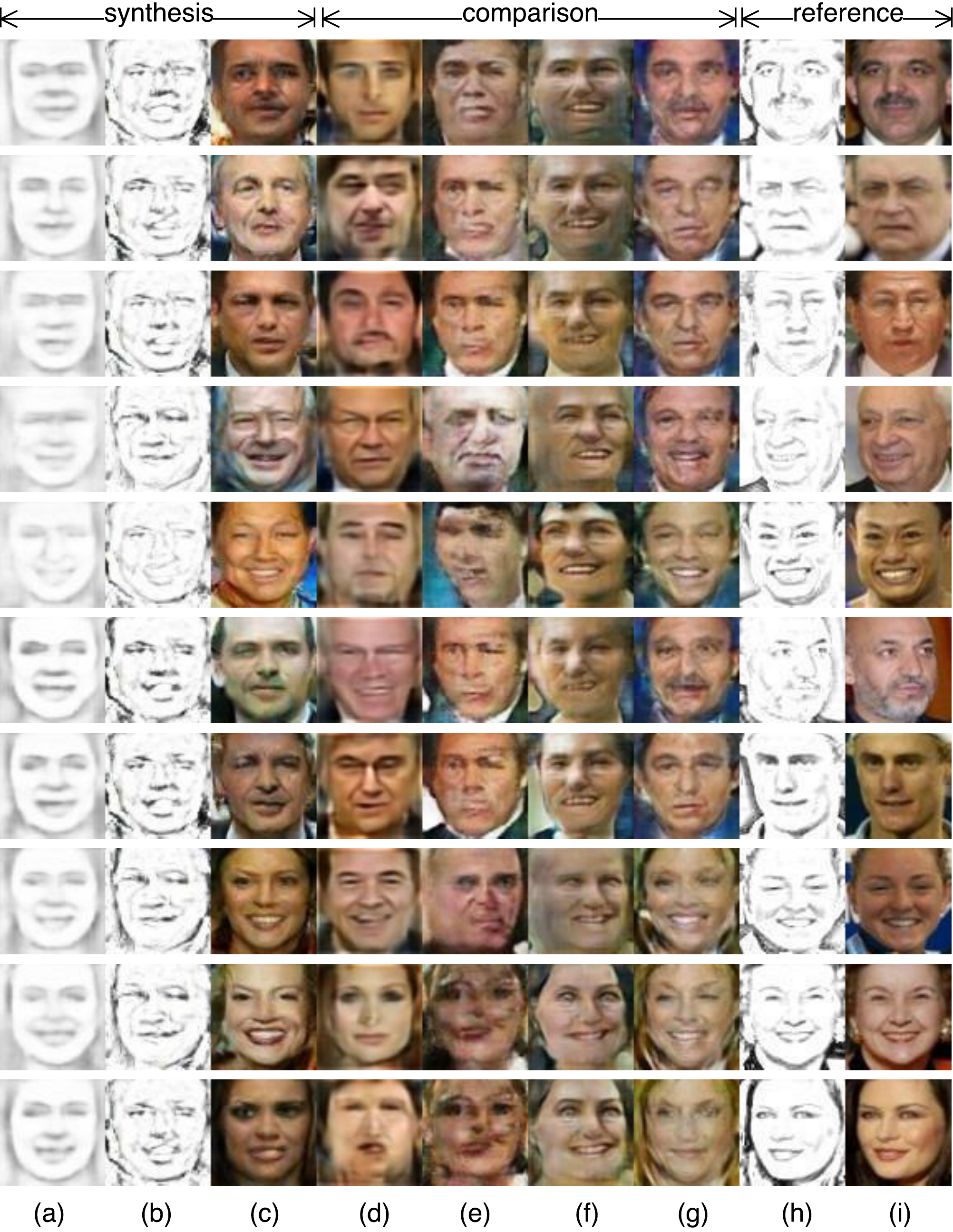}
\caption{Image reconstruction results on the LFWA dataset. (a)  Stage 1 results, (b)  Stage 2 results, (c) Stage 3 results (output from our method), (d) CVAE \cite{yan2016attribute2image}, (e) text2img \cite{reed2016generative}, (f) StackGAN \cite{zhang2016stackgan}, (g) attr2face, (h) reference sketch, (i) reference face image.}
\label{fig:LFW_result}
\end{figure*}

Sample image reconstruction results corresponding to different methods from the CelebA test set are shown in Figure~\ref{fig:CelebA_result}.  As can be seen from this figure, text2img and StackGAN methods are able to provide attribute-preserved reconstructions, but the synthesized face images are distorted and contain many artifacts.  The CVAE method is able to reconstruct the images without distortions but they are blurry.  Also, some of the attributes are difficult to see in the reconstructions from the CVAE method.  For example, hair color is hard to see in the reconstructed images.  The attr2face baseline provides reasonable reconstructions but images are distorted.  In comparison to these methods, the proposed method, as shown in (c), provides the best attribute-preserved reconstructions.  This can be seen by comparing the attributes of images in (i) with (c).  To show the improvements obtained from different stages of our method, we also show the results from Stage 1 and Stage 2 in (a) and (b), respectively.

\subsection{LFWA Dataset Results}
Images in the LFWA dataset come from the LFW dataset \cite{Huang2012a}, \cite{LFWTech}, and the corresponding attributes come from \cite{liu2015faceattributes}. This dataset contains the same 40 binary attributes as in the CelebA dataset. After preprocessing, the training and testing subsets contain 6,263 and 6,880 samples, respectively. The learning strategy for the ADAM method is the same as the one used for the CelebA dataset except that the initial learning rate is kept the same for the first 20 epochs and is dropped by 1/decay\_epoch of its previous value after every epoch which is 1/20.

Sample results corresponding to different methods on the the LFWA dataset are shown  in Figure~\ref{fig:LFW_result}.  As can be seen from the results, the CVAE method produces reconstructions which are blurry and distorted.  Attibute-conditioned GAN-based approaches such as  text2img and  StackGAN produce poor quality results with many distortions.  The attr2face baseline and the proposed method show better reconstruction compared to the other methods.  By comparing the reconstructions from our method in (c) with the images in (i) we see that the proposed method is able to reconstruct high-quality attribute-preserved face images.  Again, outputs from Stage 1 and Stage 2 of our method are shown in (a) and (b), respectively.

\subsection{CUHK Dataset Results}
\begin{figure*}[htp!]
\centering
\includegraphics[width=.8\linewidth]{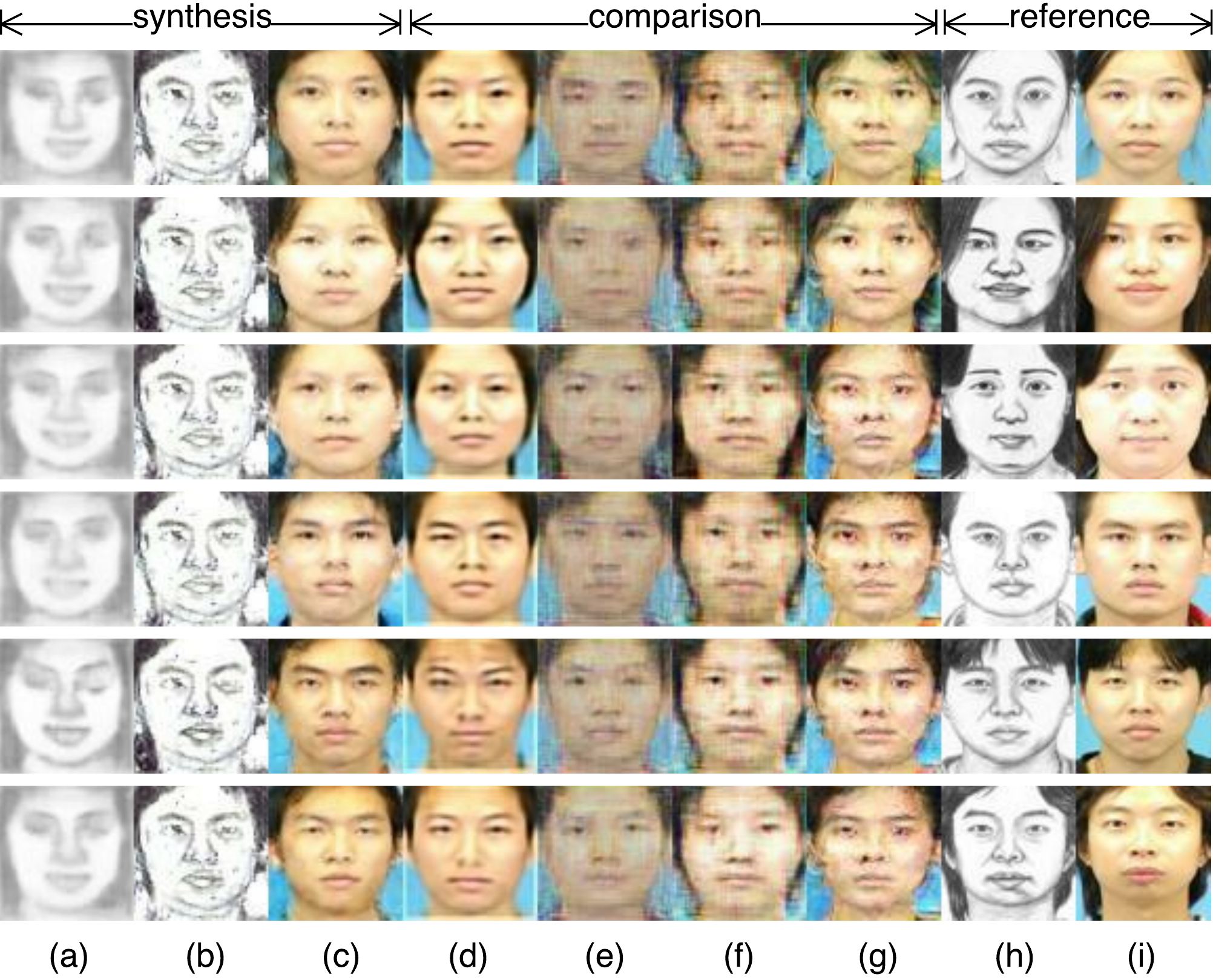}
\caption{Image reconstruction results on the CUHK dataset. (a)  Stage 1 results, (b)  Stage 2 results, (c) Stage 3 results (output from our method), (d) CVAE \cite{yan2016attribute2image}, (e) text2img \cite{reed2016generative} , (f) StackGAN \cite{zhang2016stackgan}, (g) attr2face, (h) reference sketch, (i) reference face image.}
\label{fig:cuhk_result}
\end{figure*}

Instead of using the composed sketches as was done for the experiments on the CelebA and LFWA datasets, in this section, we implemented our algorithm using real sketches and photos from the CUFS dataset \cite{wang2009face}. The CUFS dataset is a relatively small dataset. After preprocessing and data augmentation, such as flipping and rotation, we obtained 264 samples for training, and 300 samples for testing.  The batch size of 8 was used while training our network.  The other settings are kept the same as the CalebA dataset.  Since this dataset does not come with attribute annotations, we manually annotated 23 attributes on this dataset.  

Results corresponding to different methods are shown in Figure~\ref{fig:cuhk_result}.  We obtain similar results as we did in the CelebA and LFWA datasets.  The text2img, StackGAN, and attr2face methods generate images with some visual artifacts, while the CVAE method produces blurry results. In contrast, our method produces the best results and generates photo-realistic and attribute-preserved face reconstructions.

\subsection{Face Synthesis}
\begin{figure*}[htp!]
\centering
\includegraphics[width=1\linewidth]{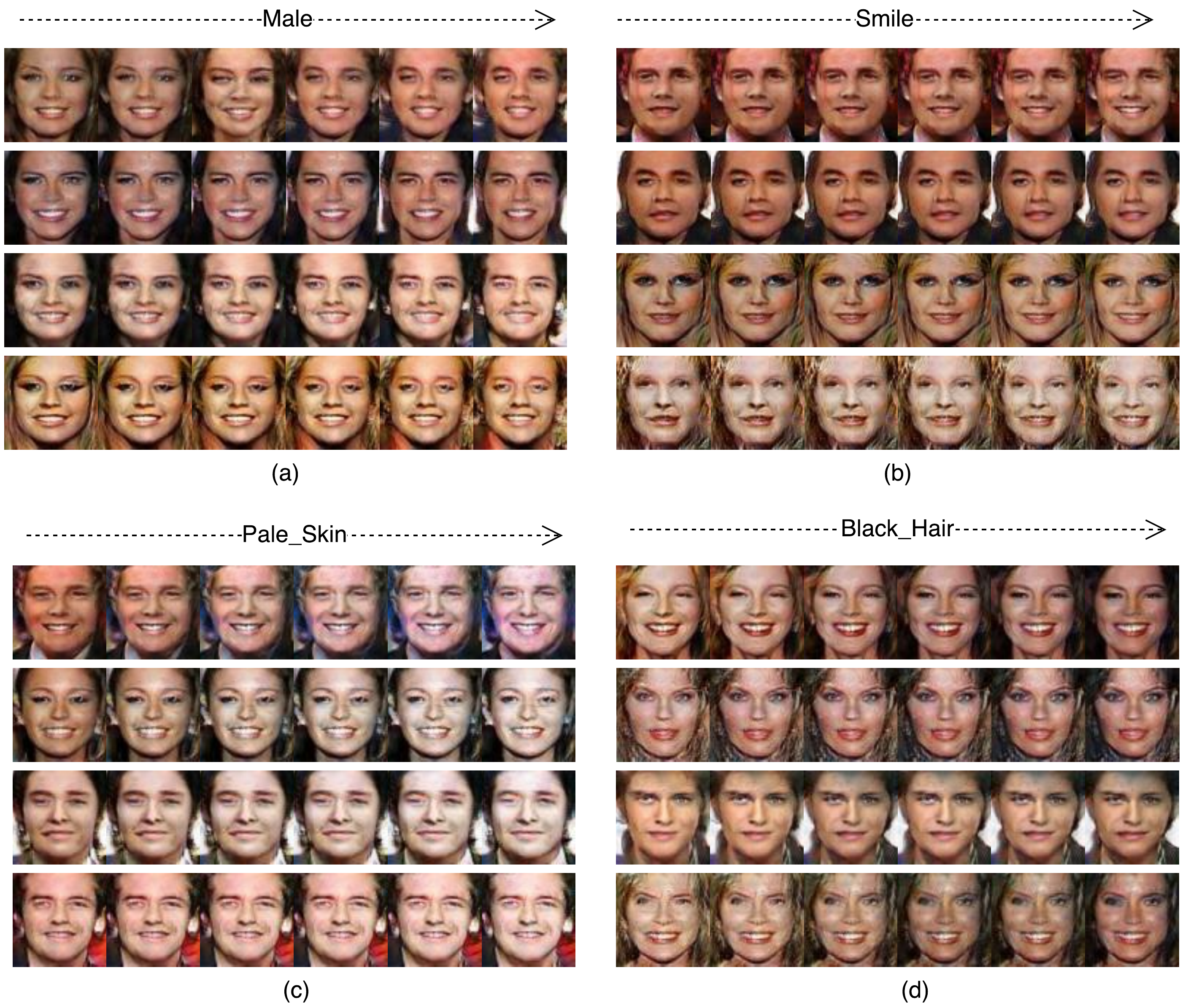}
\caption{Sample image synthesis on CelebA when attributes are changed while the noise vector is kept frozen.  (a) Female to male. (b) Neutral to smile. (c) Original skin tone to pale skin tone. (d) Original hair color to black hair color.}
\label{fig:CelebA_progression}
\end{figure*}

\begin{figure*}[htp!]
\centering
\includegraphics[width=1\linewidth]{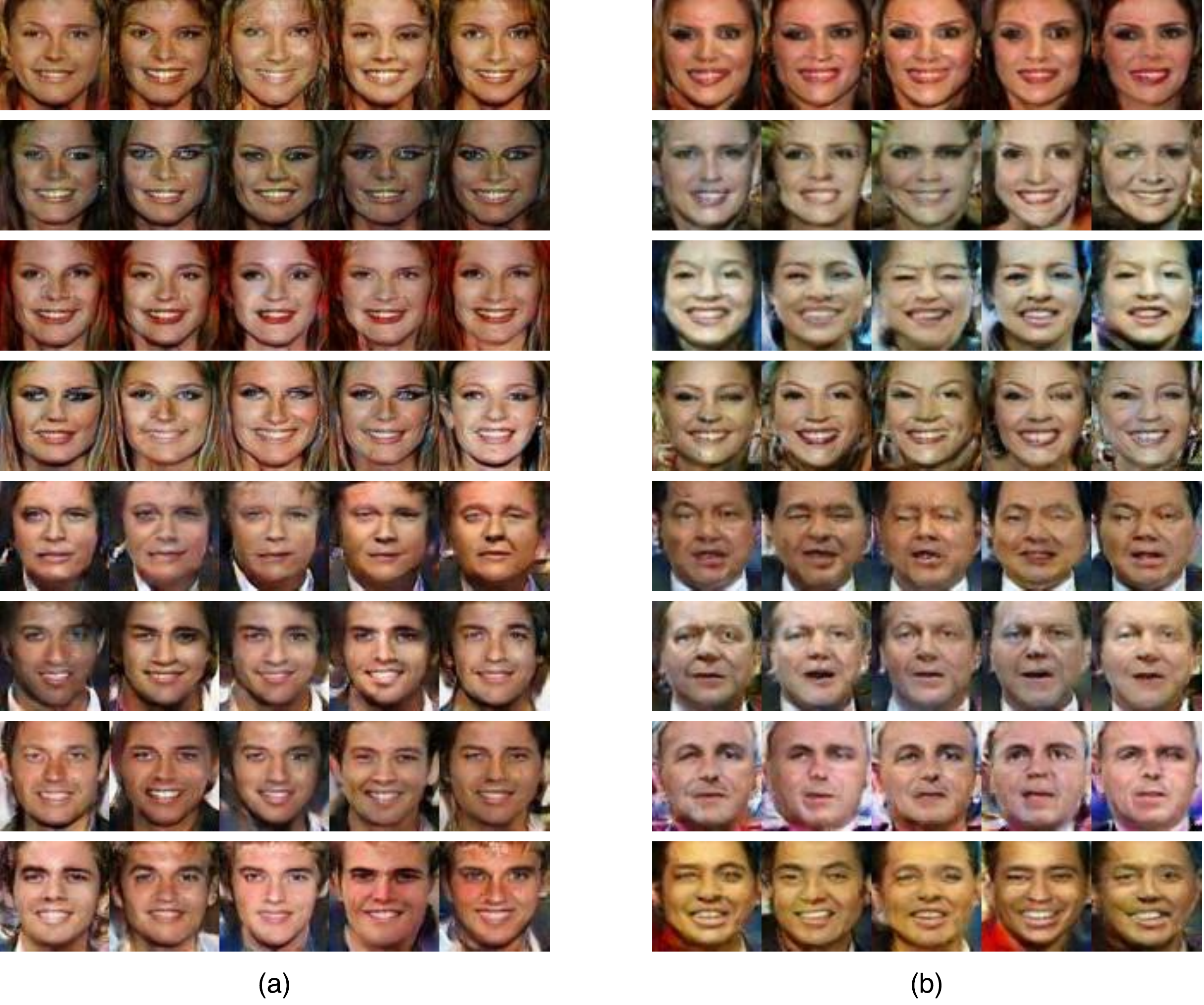}
\caption{Sample image synthesis results on (a) CelebA, and (b) LFWA when attributes are kept frozen while the noise vector is changed according to $n\sim N(0,1)$. Note that the identity changes as we vary the noise vector but he attributes stay the same on the reconstructed images.}
\label{fig:random_noise}
\end{figure*}

\begin{table*}[t]
\centering
\caption{Quantitative results corresponding to different methods.The Inception Score and Attribute $L_{2}$ measure are used to compare the performance of different methods.}
\label{tab: quantitative result}
\begin{tabular}{|c|c|c|c|c|c|c|}
\hline Metric  & Dataset & text2img \cite{reed2016generative} & StackGAN \cite{zhang2016stackgan} & CVAE \cite{yan2016attribute2image} & attr2face & Attribute2Sketch2Face	 \\
\hline 
\multirow{3}{*}{Inception Score} 
& CelebA & $1.486 \pm 0.016$ & $1.517 \pm 0.014$ & $1.275 \pm 0.005$ & $1.524 \pm 0.010$ & $\boldsymbol{1.545 \pm 0.011}$ \\
& LFW & $1.510 \pm 0.020$ & $1.589 \pm 0.018$ & $1.482 \pm 0.017$ & $1.592 \pm 0.081$ & $\boldsymbol{1.617 \pm 0.025}$ \\
& CUHK & $1.269 \pm 0.049$ & $1.349 \pm 0.114$ & $1.119 \pm 0.027$ & $1.345 \pm 0.131$ & $\boldsymbol{1.417 \pm 0.040}$ \\ 
\hline 
\multirow{3}{*}{Attribute $L_{2}$} 
& CelebA & $0.104 \pm 0.024$  & $0.091 \mp 0.021$  & $0.080\pm 0.042$ & $0.081 \pm 0.023$ & $\boldsymbol{0.079 \pm 0.022}$ \\
& LFW & $0.093 \pm 0.027 $ & $0.085\pm 0.029$ & $0.086\pm 0.019$ & $0.073 \pm 0.035$ & $\boldsymbol{0.069 \pm 0.034}$ \\
& CUHK & $0.610\pm 0.018$ & $0.063 \pm 0.019$ & $0.064 \pm 0.020$ & $0.062 \pm 0.020 $ & $\boldsymbol{0.058 \pm 0.021}$ \\
\hline 
\end{tabular} \end{table*}

In this section, we show the image synthesis capability of our network by manipulating the input attribute and noise vectors.  Note that, the testing phase of our network takes attribute vector and noise as inputs and produces face reconstruction as the output. In the first set of experiments with image synthesis, we keep the random noise vector the same, i.e. $n\sim N(0,1)$ and change the attribute weights corresponding to a particular attribute as follows: $[-1, -0.1, 0.1, 0.4, 0.7, 1]$.  The corresponding results on the CelebA dataset are shown in Figure~\ref{fig:CelebA_progression}.  From this figure, we can see that when we give higher weights to a certain attribute, the corresponding appearance changes.  For example, one can 
 synthesize an image with a different gender by changing the weights corresponding to the gender attribute as shown in Figure~\ref{fig:CelebA_progression}(a).  Each row shows the progression of gender change as the attribute weights are changed from -1 to 1 as described above.  Similarly, figures (b), (c) and (d) show the synthesis results when a neutral face image is transformed into a smily face image, skin tones are changed to pale skin tone, and hair colors are changed to black, respectively.    It is interesting to see that when the attribute weights other than the gender attribute are changed, the identity of the person does not change.  Only the attributes change.

In the second set of experiments, we keep the input attribute vector frozen but now change the noise vector by inputing different realizations of $n\sim N(0,1)$.   Sample results corresponding to this experiment are shown in Figure~\ref{fig:random_noise}(a) and (b) using the CelebA and LFWA datasets, respectively.  Each column shows how the output changes as we change the noise vector.  Different subjects are shown in different rows.  It is interesting to note that, as we change the noise vector, attributes stay the same while the identity changes.  This can be clearly seen by comparing the reconstructions in each row.

\subsection{Quantitative Results}
In addition to the qualitative results presented in Figures~\ref{fig:CelebA_result}, \ref{fig:LFW_result} and \ref{fig:cuhk_result}, we present quantitative comparisons based on the Inception Score \cite{salimans2016improved} and Attribute $L_{2}$-norm.    The inception scores are used to evaluate the realism and diversity of the generated samples and has been used before to evaluate the performance of deep generative methods \cite{bao2017cvae}, \cite{zhang2016stackgan}.  Attribute $L_{2}$-norm is used to compare the quality of attributes corresponding to different images. We extract the attributes from the synthesized images as well as the reference image using the MOON attribute prediction method \cite{rudd2016moon}.  Once the attributes are extracted, we simply take the $L_{2}$-norm of the difference between the attributes as follows
\begin{equation}
\text{Attribute  } L_{2}=\|\hat{a}_{ref}-\hat{a}_{synth}\|_{2},
\end{equation}
where $\hat{a}_{ref}$ and $\hat{a}_{synth}$ are the 23 extracted attributes from the reference image and the synthesized image, respectively.  Note that higher values of the Inception Score  and lower values of the Attribute $L_{2}$ measure imply the better performance.   The  quantitive results corresponding to different methods on the CalebA, LFW and CUHK datasets are shown in Table~\ref{tab: quantitative result}.  Results are evaluated on the test splits of the corresponding dataset and the average performance along with the standard deviation are reported in Table~\ref{tab: quantitative result}.

As can be seen from this table, the proposed Attribute2Sketch2Face method produces the highest inception scores implying that the images generated by our method are more realistic than the ones generated by other methods.  Furthermore, our method produces the lowest Attribute $L_{2}$ scores.  This implies that our method is able to generate attribute-preserved images better than the other compared methods.   This can be clearly seen by comparing the images synthesized by different methods in Figures~\ref{fig:CelebA_result}, \ref{fig:LFW_result} and \ref{fig:cuhk_result}.

\section{Conclusion}\label{sec:con}
We presented a novel deep generative framework for reconstructing face images from visual attributes.  Our method makes use of an intermediate representation to generate photo realistic images.  The training part of our method consists of three stages - A2S, S2S and S2F.  The A2S stage is based on the CVAE model while the S2S and S2F stages are based on GANs.  Novel UNet-based generators are proposed for the S2S and S2F stages.    Various experiments on three publicly available datasets show the significance of the proposed three-stage synthesis framework.  In addition, an ablation study was conducted to show the importance of different components of our network.   Various experiments showed that the proposed method is able to generate high-quality images and achieves significant improvements over the state-of-the-art methods.

\begin{acknowledgements}
This research is based upon work supported by the Office
of  the  Director  of  National  Intelligence  (ODNI),  Intelligence  Advanced  Research  Projects  Activity  (IARPA),  via
IARPA  R\&D  Contract  No.  2014-14071600012.  The  views
and  conclusions  contained  herein  are  those  of  the  authors
and should not be interpreted as necessarily representing the
official policies or endorsements, either expressed or implied,
of  the  ODNI,  IARPA,  or  the  U.S.  Government.  The  U.S.
Government is authorized to reproduce and distribute reprints
for  Governmental  purposes  notwithstanding  any  copyright
annotation  thereon.
\end{acknowledgements}

\bibliographystyle{spmpsci}      % mathematics and physical sciences
\bibliography{egbib}   % name your BibTeX data base

\begin{thebibliography}{10}
\providecommand{\url}[1]{{#1}}
\providecommand{\urlprefix}{URL }
\expandafter\ifx\csname urlstyle\endcsname\relax
  \providecommand{\doi}[1]{DOI~\discretionary{}{}{}#1}\else
  \providecommand{\doi}{DOI~\discretionary{}{}{}\begingroup
  \urlstyle{rm}\Url}\fi

\bibitem{arjovsky2017towards}
Arjovsky, M., Bottou, L.: Towards principled methods for training generative
  adversarial networks.
\newblock arXiv preprint arXiv:1701.04862  (2017)

\bibitem{arjovsky2017wasserstein}
Arjovsky, M., Chintala, S., Bottou, L.: Wasserstein gan.
\newblock arXiv preprint arXiv:1701.07875  (2017)

\bibitem{bao2017cvae}
Bao, J., Chen, D., Wen, F., Li, H., Hua, G.: Cvae-gan: Fine-grained image
  generation through asymmetric training.
\newblock arXiv preprint arXiv:1703.10155  (2017)

\bibitem{berthelot2017began}
Berthelot, D., Schumm, T., Metz, L.: Began: Boundary equilibrium generative
  adversarial networks.
\newblock arXiv preprint arXiv:1703.10717  (2017)

\bibitem{che2016mode}
Che, T., Li, Y., Jacob, A.P., Bengio, Y., Li, W.: Mode regularized generative
  adversarial networks.
\newblock arXiv preprint arXiv:1612.02136  (2016)

\bibitem{softbio}
Dantcheva, A., Elia, P., Ross, A.: What else does your biometric data reveal? a
  survey on soft biometrics.
\newblock IEEE Transactions on Information Forensics and Security
  \textbf{11}(3), 441--467 (2016)

\bibitem{denton2015deep}
Denton, E.L., Chintala, S., Fergus, R., et~al.: Deep generative image models
  using a￼ laplacian pyramid of adversarial networks.
\newblock In: Advances in neural information processing systems, pp. 1486--1494
  (2015)

\bibitem{dosovitskiy2017learning}
Dosovitskiy, A., Springenberg, J.T., Tatarchenko, M., Brox, T.: Learning to
  generate chairs, tables and cars with convolutional networks.
\newblock IEEE transactions on pattern analysis and machine intelligence
  \textbf{39}(4), 692--705 (2017)

\bibitem{gauthier2014conditional}
Gauthier, J.: Conditional generative adversarial nets for convolutional face
  generation

\bibitem{goodfellow2014generative}
Goodfellow, I., Pouget-Abadie, J., Mirza, M., Xu, B., Warde-Farley, D., Ozair,
  S., Courville, A., Bengio, Y.: Generative adversarial nets.
\newblock In: Advances in neural information processing systems, pp. 2672--2680
  (2014)

\bibitem{huang2017densely}
Huang, G., Liu, Z., van~der Maaten, L., Weinberger, K.Q.: Densely connected
  convolutional networks.
\newblock In: Proceedings of the IEEE Conference on Computer Vision and Pattern
  Recognition (2017)

\bibitem{huang2016densely}
Huang, G., Liu, Z., Weinberger, K.Q., van~der Maaten, L.: Densely connected
  convolutional networks.
\newblock arXiv preprint arXiv:1608.06993  (2016)

\bibitem{Huang2012a}
Huang, G.B., Mattar, M., Lee, H., Learned-Miller, E.: Learning to align from
  scratch.
\newblock In: NIPS (2012)

\bibitem{LFWTech}
Huang, G.B., Ramesh, M., Berg, T., Learned-Miller, E.: Labeled faces in the
  wild: A database for studying face recognition in unconstrained environments.
\newblock Tech. Rep. 07-49, University of Massachusetts, Amherst (2007)

\bibitem{isola2016image}
Isola, P., Zhu, J.Y., Zhou, T., Efros, A.A.: Image-to-image translation with
  conditional adversarial networks.
\newblock arXiv preprint arXiv:1611.07004  (2016)

\bibitem{jegou2017one}
J{\'e}gou, S., Drozdzal, M., Vazquez, D., Romero, A., Bengio, Y.: The one
  hundred layers tiramisu: Fully convolutional densenets for semantic
  segmentation.
\newblock In: Computer Vision and Pattern Recognition Workshops (CVPRW), 2017
  IEEE Conference on, pp. 1175--1183. IEEE (2017)

\bibitem{johnson2016perceptual}
Johnson, J., Alahi, A., Fei-Fei, L.: Perceptual losses for real-time style
  transfer and super-resolution.
\newblock In: European Conference on Computer Vision, pp. 694--711. Springer
  (2016)

\bibitem{adam_opt}
Kingma, D., Ba, J.: Adam: A method for stochastic optimization.
\newblock arXiv preprint arXiv:1412.6980  (2014)

\bibitem{kingma2013auto}
Kingma, D.P., Welling, M.: Auto-encoding variational bayes.
\newblock arXiv preprint arXiv:1312.6114  (2013)

\bibitem{kumar2008facetracer}
Kumar, N., Belhumeur, P., Nayar, S.: Facetracer: A search engine for large
  collections of images with faces.
\newblock In: European conference on computer vision, pp. 340--353. Springer
  (2008)

\bibitem{kumar_ttributes}
Kumar, N., Berg, A., Belhumeur, P.N., Nayar, S.: Describable visual attributes
  for face verification and image search.
\newblock IEEE Transactions on Pattern Analysis and Machine Intelligence
  \textbf{33}(10), 1962--1977 (2011)

\bibitem{larochelle2011neural}
Larochelle, H., Murray, I.: The neural autoregressive distribution estimator.
\newblock In: Proceedings of the Fourteenth International Conference on
  Artificial Intelligence and Statistics, pp. 29--37 (2011)

\bibitem{larsen2015autoencoding}
Larsen, A.B.L., S{\o}nderby, S.K., Larochelle, H., Winther, O.: Autoencoding
  beyond pixels using a learned similarity metric.
\newblock arXiv preprint arXiv:1512.09300  (2015)

\bibitem{DeepAtt}
Liu, Z., Luo, P., Wang, X., Tang, X.: Deep learning face attributes in the
  wild.
\newblock In: 2015 IEEE International Conference on Computer Vision (ICCV), pp.
  3730--3738 (2015)

\bibitem{liu2015faceattributes}
Liu, Z., Luo, P., Wang, X., Tang, X.: Deep learning face attributes in the
  wild.
\newblock In: Proceedings of International Conference on Computer Vision (ICCV)
  (2015)

\bibitem{mao2016image}
Mao, X.J., Shen, C., Yang, Y.B.: Image denoising using very deep fully
  convolutional encoder-decoder networks with symmetric skip connections.
\newblock arXiv preprint  (2016)

\bibitem{metz2016unrolled}
Metz, L., Poole, B., Pfau, D., Sohl-Dickstein, J.: Unrolled generative
  adversarial networks.
\newblock arXiv preprint arXiv:1611.02163  (2016)

\bibitem{mirza2014conditional}
Mirza, M., Osindero, S.: Conditional generative adversarial nets.
\newblock arXiv preprint arXiv:1411.1784  (2014)

\bibitem{odena2016conditional}
Odena, A., Olah, C., Shlens, J.: Conditional image synthesis with auxiliary
  classifier gans.
\newblock arXiv preprint arXiv:1610.09585  (2016)

\bibitem{radford2015unsupervised}
Radford, A., Metz, L., Chintala, S.: Unsupervised representation learning with
  deep convolutional generative adversarial networks.
\newblock arXiv preprint arXiv:1511.06434  (2015)

\bibitem{reed2016generative}
Reed, S., Akata, Z., Yan, X., Logeswaran, L., Schiele, B., Lee, H.: Generative
  adversarial text to image synthesis.
\newblock In: M.F. Balcan, K.Q. Weinberger (eds.) Proceedings of The 33rd
  International Conference on Machine Learning, \emph{Proceedings of Machine
  Learning Research}, vol.~48, pp. 1060--1069. New York, New York, USA (2016)

\bibitem{rezende2014stochastic}
Rezende, D.J., Mohamed, S., Wierstra, D.: Stochastic backpropagation and
  approximate inference in deep generative models.
\newblock arXiv preprint arXiv:1401.4082  (2014)

\bibitem{ronneberger2015u}
Ronneberger, O., Fischer, P., Brox, T.: U-net: Convolutional networks for
  biomedical image segmentation.
\newblock In: International Conference on Medical Image Computing and
  Computer-Assisted Intervention, pp. 234--241. Springer (2015)

\bibitem{rudd2016moon}
Rudd, E.M., G{\"u}nther, M., Boult, T.E.: Moon: A mixed objective optimization
  network for the recognition of facial attributes.
\newblock In: European Conference on Computer Vision, pp. 19--35. Springer
  (2016)

\bibitem{salimans2016improved}
Salimans, T., Goodfellow, I., Zaremba, W., Cheung, V., Radford, A., Chen, X.:
  Improved techniques for training gans.
\newblock In: Advances in Neural Information Processing Systems, pp. 2234--2242
  (2016)

\bibitem{simonyan2014very}
Simonyan, K., Zisserman, A.: Very deep convolutional networks for large-scale
  image recognition.
\newblock arXiv preprint arXiv:1409.1556  (2014)

\bibitem{sohn2015learning}
Sohn, K., Lee, H., Yan, X.: Learning structured output representation using
  deep conditional generative models.
\newblock In: Advances in Neural Information Processing Systems, pp. 3483--3491
  (2015)

\bibitem{disentangled}
Tran, L., Yin, X., Liu, X.: Disentangled representation learning gan for
  pose-invariant face recognition.
\newblock In: In Proceeding of IEEE Computer Vision and Pattern Recognition.
  Honolulu, HI (2017)

\bibitem{wang2009face}
Wang, X., Tang, X.: Face photo-sketch synthesis and recognition.
\newblock IEEE Transactions on Pattern Analysis and Machine Intelligence
  \textbf{31}(11), 1955--1967 (2009)

\bibitem{yan2016attribute2image}
Yan, X., Yang, J., Sohn, K., Lee, H.: Attribute2image: Conditional image
  generation from visual attributes.
\newblock In: European Conference on Computer Vision, pp. 776--791. Springer
  (2016)

\bibitem{zhang2016stackgan}
Zhang, H., Xu, T., Li, H., Zhang, S., Huang, X., Wang, X., Metaxas, D.:
  Stackgan: Text to photo-realistic image synthesis with stacked generative
  adversarial networks.
\newblock arXiv preprint arXiv:1612.03242  (2016)

\bibitem{zhang2016joint}
Zhang, K., Zhang, Z., Li, Z., Qiao, Y.: Joint face detection and alignment
  using multitask cascaded convolutional networks.
\newblock IEEE Signal Processing Letters \textbf{23}(10), 1499--1503 (2016)

\bibitem{zhang2014panda}
Zhang, N., Paluri, M., Ranzato, M., Darrell, T., Bourdev, L.: Panda: Pose
  aligned networks for deep attribute modeling.
\newblock In: Proceedings of the IEEE conference on computer vision and pattern
  recognition, pp. 1637--1644 (2014)

\bibitem{zhu2017unpaired}
Zhu, J.Y., Park, T., Isola, P., Efros, A.A.: Unpaired image-to-image
  translation using cycle-consistent adversarial networks.
\newblock arXiv preprint arXiv:1703.10593  (2017)

\end{thebibliography}

\end{document}